\documentclass[letterpaper, 10 pt, conference]{ieeeconf}  

\IEEEoverridecommandlockouts  
\usepackage{amsmath}
\usepackage{amssymb}
\usepackage[nospace]{cite}
\usepackage{url}
\usepackage{xcolor}
\usepackage{fancyhdr}
\usepackage[hidelinks]{hyperref}

\usepackage{adjustbox}
\usepackage{graphicx}
\usepackage{algorithm}
\usepackage[noend]{algpseudocode}
\usepackage{tensor}

\usepackage{caption}
\usepackage{subcaption}

\usepackage[utf8]{inputenc}
\usepackage[OT1]{fontenc}
\usepackage[final]{microtype}

\usepackage{booktabs}

\usepackage[binary-units]{siunitx}
\usepackage{gensymb} 
\usepackage{multirow}
\usepackage{lipsum}
\usepackage[font=scriptsize]{caption} 

\definecolor{darkblue}{rgb}{0, .0, 5}

\title{\LARGE \bf
LiDAR-guided object search and detection in Subterranean Environments
}

\author{Manthan Patel, Gabriel Waibel, Shehryar Khattak, Marco Hutter
\thanks{This work was supported in part by the ETH RobotX student fellowship}
\thanks{The authors are with the Robotic Systems Lab, ETH Z\"urich}
\thanks{Correspondence email: \tt\small patelm@ethz.ch}
}

\begin{document}

\maketitle
\thispagestyle{empty}
\pagestyle{empty}

\begin{abstract}

Detecting objects of interest, such as human survivors, safety equipment, and structure access points, is critical to any search-and-rescue operation. Robots deployed for such time-sensitive efforts rely on their onboard sensors to perform their designated tasks. However, as disaster response operations are predominantly conducted under perceptually degraded conditions, commonly utilized sensors such as visual cameras and LiDARs suffer in terms of performance degradation. In response, this work presents a method that utilizes the complementary nature of vision and depth sensors to leverage multi-modal information to aid object detection at longer distances. In particular, depth and intensity values from sparse LiDAR returns are used to generate proposals for objects present in the environment. These proposals are then utilized by a Pan-Tilt-Zoom (PTZ) camera system to perform a directed search by adjusting its pose and zoom level for performing object detection and classification in difficult environments. The proposed work has been thoroughly verified using an ANYmal quadruped robot in underground settings and on datasets collected during the DARPA Subterranean Challenge finals.
\end{abstract}


\section{Introduction}\label{sec:intro}
Rapid advancement of robotics systems over the past decade have facilitated their application towards time-sensitive and mission-critical operations, such as search-and-rescue~\cite{bogue2019disaster,siegwart2015legged}, disaster response~\cite{thakur2019nuclear,visionThermal} and infrastructure inspection~\cite{roboticinspectionsurvey,changeDetetion}, across complex environments and under difficult operational conditions.
In response, field-ready robotic deployment in challenging scenarios has recently become an area of interest for robotics researchers and the wider stakeholder audience, as showcased by the recently concluded DARPA Subterranean (SubT) Challenge~\cite{tranzatto2022team}. A key performance indicator for the SubT challenge, in particular, and for search-and-rescue missions, in general, is to detect and identify objects of interest while exploring the target areas. Human survivors, human identifiers (such as clothes, helmets, and backpacks), safety equipment (such as fire extinguishers and tools), and environment access points (such as doors and ducts) constitute examples of the vital objects that need to be detected in a time-efficient manner using on-board sensors of the robots during critical tasks. 

\begin{figure}[ht!]
    \centering
    \includegraphics[width=\columnwidth]{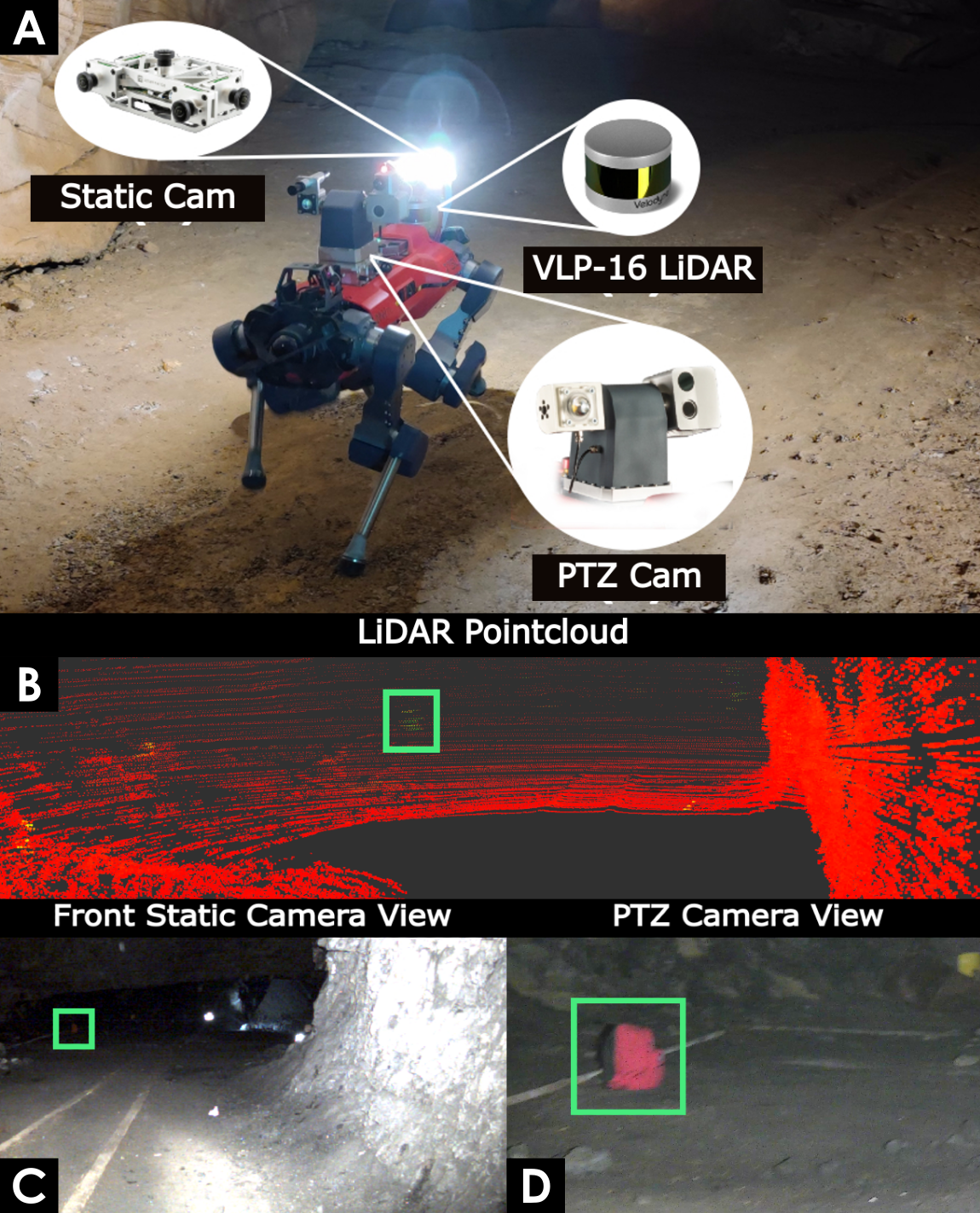}
    \caption{(A) Instance of autonomous exploration mission conducted at the Salt-Peter Cave, Kentucky, USA using the ANYmal-C robot, with zoomed-in views showing relevant onboard sensors. Far-away objects cannot be detected by static cameras alone due to limited illumination (C). Utilizing the proposed method, object proposals are generated from LiDAR (B) to direct the Pan-Tilt-Zoom camera to perform detection, resulting in the backpack (D) being detected at \SI{10}{\meter} as compared to \SI{3}{\meter} when using only a static camera.}
    \label{fig:cover}    
\vspace{-5ex}
\end{figure}

For object detection, visual cameras have been the sensor of choice due to their cost-effective, lightweight, and power-efficient nature. However, in disaster response scenarios, poor illumination and the presence of obscurants, such as dust, smoke, and fog, severely degrade the camera performance and effectively limits the range of object detection to a couple of meters. Furthermore, rigidly mounted (static) cameras provide only a limited observation of the environment and depend on the robot pose to be such that the object is within their Field-of-View (FoV) to be detected. Articulated cameras, such as the Pan-Tilt-Zoom (PTZ) camera shown in Figure~\ref{fig:cover}, can independently orient themselves and change their zoom level to obtain better observation of far-away objects. Nevertheless, given the large number of combinations of orientations and zoom levels required to fully observe the surrounding environment, it is typically not feasible to perform complete coverage without impacting the speed of environment exploration. In contrast, LiDAR sensors provide 360$\degree$ observation, depth measurements at long-range, and remain unaffected by scene illumination, providing an alternate sensor choice for object detection. However, the sparse nature and low fidelity of LiDAR data compared to visual data make accurate object detection difficult.

Motivated by the discussion above, this work presents a method that utilizes the complementary nature of camera and LiDAR data to facilitate object detection at long ranges. In particular, depth and intensity values from sparse LiDAR returns are used to detect and generate location proposals for the objects present in the environment. These location proposals are then used by a PTZ camera system to perform a directed search by adjusting its orientation and zoom level to perform object detection and classification in difficult environments at long ranges. The performance and applicability of the proposed method is thoroughly evaluated on data collected by an ANYmal-C quadruped robot during field deployments conducted in challenging underground settings, including the SubT Challenge finals event consisting of an underground urban environment, a cave network, and a tunnel system.




\section{Related Work}\label{sec:related}
\label{sec:relatedwork}
Visual cameras have been the preferred sensor choice for object detection due to having rich scene information including texture and context. Especially with the emergence of Convolutional Neural Network (CNN) based object detection approaches such as YOLO~\cite{YOLO}, SSD~\cite{SSD}, faster R-CNN~\cite{Fasterrcnn}, on-par human-level performance has been achieved. Moreover, recent approaches such as Mask R-CNN \cite{mask-rcnn}, DetectoRS \cite{DetectoRs} are able to perform instance segmentation in which each pixel of the image is assigned a class label and an instance label. However, due to the absence of depth information, localizing the detected objects in 3D environment remains a challenge. This has motivated the teams participating in the DARPA SubT challenge to use LiDAR scans for localizing the detected objects. Team CERBERUS~\cite{dang2020autonomous} made use of a YOLO architecture trained to include competition-specific objects for detection. The 3D location of the object in world coordinates is obtained by projecting the bounding box into the robot occupancy map built using the LiDAR scans.
Other teams also made use of similar approaches utilizing both camera and LiDAR data~\cite{Costar-OD, CSIRO}. A common problem reported by all teams was the reduced object detection range using only visual cameras due to poor illumination in complex underground environments.

LiDAR-based 3D object detection methods which make use of CNNs and operate on point clouds (Point R-CNN \cite{Point-rcnn}) or voxel-based representation (Voxel R-CNN \cite{Voxel-RCNN}) have also gained popularity. While these approaches are well suited for detecting and localizing objects like vehicles and pedestrians in a structured environment like that of a self-driving vehicle, they are not well suited for detecting highly specific objects in an unstructured environment as required in our case. Thus, we propose to use LiDAR and a PTZ camera in a coupled manner to improve the object detection range. In particular, we propose to use LiDAR scans to generate object proposals by performing clustering based on LiDAR intensity and depth difference. These clusters are then scanned by the PTZ camera and classified using a CNN-based object detection model.
Existing methods have performed object segmentation and clustering using sparse LiDAR scans, with a simple clustering approach based on the Euclidean distance proposed in~\cite{Euclidean-clustering}. The approach operates directly on the 3D point clouds and introduces a radially bounded nearest neighbor algorithm for clustering which is able to handle outliers as opposed to a 'k' nearest neighbor clustering\cite{Knn-clustering}. This approach was further extended in \cite{eucliden-clustering-normal} to work in real-time on a continuous stream of data. Methods operating directly on unordered point clouds are relatively slow due to the expensive nearest neighbor search queries. Thus, for speed-up, approaches choose to operate on range images generated from point clouds instead.
Performing computations on range images have the advantages of exploitable neighborhood relations and the reduction of redundant points to a single representative pixel in the image. In~\cite{PRBonn1}, the authors propose to use the depth angle for clustering on range images. In another clustering approach, Scan-Line-Run (SLR)~\cite{scan-line-clustering}, the authors propose to modify the two-run connected component labeling technique for binary images~\cite{SLbinary} and apply it for clustering the range images. In recent work~\cite{fastLidar}, the authors extend the depth-angle-based clustering approach of~\cite{PRBonn1} to make it robust to instance over-segmentation by introducing additional sparse connections in the range image, termed map connections.






\section{Proposed Method}\label{sec:method}
\label{sec:methodology}
\begin{figure}
    \centering
    \includegraphics[width=\columnwidth]{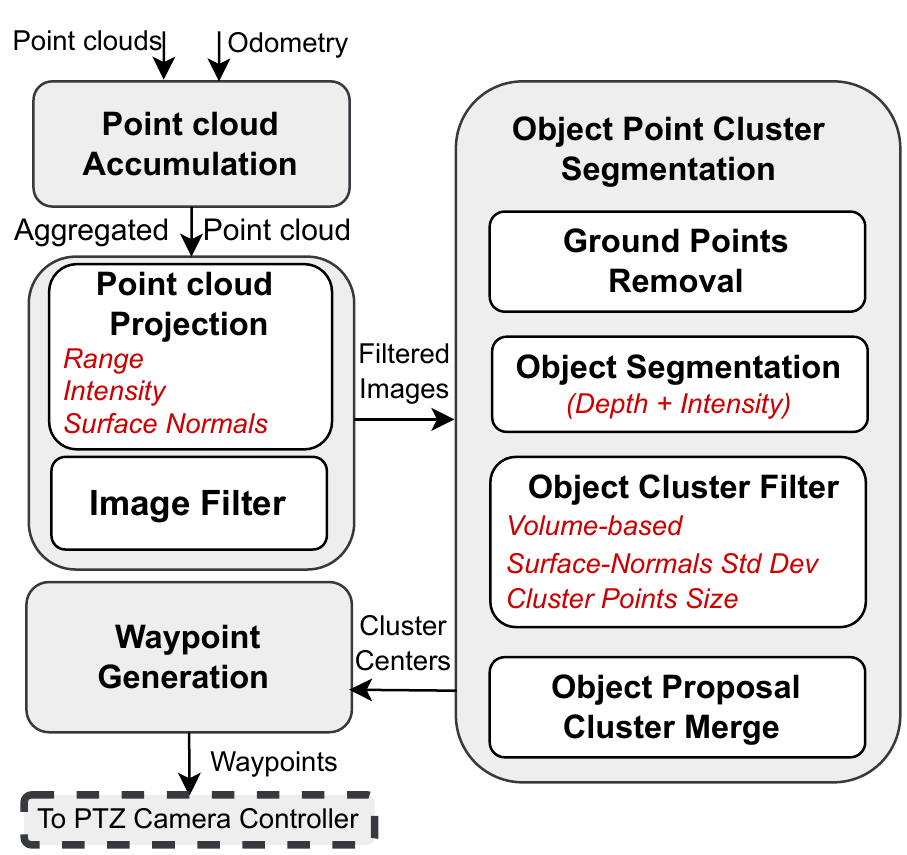}
    \caption{An overview of the proposed method.}
    \label{fig:pipeline}
\vspace{-3ex}
\end{figure}

To aid the camera object detection and classification process, especially in challenging and visually-degraded environments, this work proposes to utilize LiDAR data to generate object proposals at longer distances. In addition to utilizing depth data, this work uses auxiliary LiDAR data, such as intensity return information, to distinguish and segment objects from the environment. An overview of the proposed approach is presented in Figure~\ref{fig:pipeline}, with each component detailed below: 

\subsection{Point cloud Accumulation}\label{sec:supercloud}
To facilitate object detection from sparse LiDAR scans (Figure~\ref{fig:images}A), such as that obtained from low-cost LiDARs with fewer beams like the Velodyne VLP-16, multiple LiDARs scans are accumulated to generate a dense representation (Figure~\ref{fig:images}B). To accumulate LiDAR scans, a sliding window of size \(N_c\) is maintained.
 Upon query at a defined frequency \(f\) to generate a dense representation, all point clouds in the sliding window are transformed to the latest LiDAR frame using the robot odometry information. Furthermore, in addition to time, only point clouds satisfying a minimum motion criterion of translation \(d_{min}\) or rotation \(\theta_{min}\) are added to the sliding window to ensure good coverage and dense representation of the environment.

\subsection{Point cloud Projection}\label{sec:image}
The accumulated point cloud is converted to an ordered image representation, with rows and columns corresponding to elevation and azimuth angles respectively, to enable fast point-wise information lookup and to facilitate the application of neighborhood-based image processing operations. In addition to point coordinates, a range, a LiDAR intensity return, and a surface normals image are generated.
For all images, a fixed resolution of $180$ rows and $1200$ columns, corresponding to a vertical and horizontal angular coverage of $60\degree$ and $360\degree$ respectively, is used. In the case of accumulated LiDAR scan, as the vertical resolution covered is variable and discrete, the generated images can contain empty rows, seen as black lines in Figure~\ref{fig:images}C. To fill in missing data from neighboring scan lines, bilinear interpolation~\cite{VPR} is performed. To reduce the effect of noise and aliasing, the result is smoothed using a Gaussian kernel. The resulting filtered range and intensity images are shown in Figure~\ref{fig:images}D and \ref{fig:images}E, respectively.

In addition to depth and intensity differences, surface normals distribution is used to detect and segment objects from the environment background surface. Characteristically the surface normals distribution of an object has a higher standard deviation due to its irregular surface compared to a typical environment surface, such as a wall or floor, which commonly has a consistent surface normals distribution. To calculate a surface normals image in a computationally efficient manner, the ordered structure of the image projection is exploited, and vector cross-products between neighboring points are calculated to compute a mean surface normal for all points in the accumulated point cloud~\cite{li2019net}. The computed range, intensity, and surface normals images are then used for object clustering in the next step.

\subsection{Point Clustering for Object Segmentation} \label{sec:clustering}
For segmentation of objects from the environment, points belonging to the same object are clustered using the range, intensity, and surface normals information. First, using range image information, points belonging to the ground are removed using the approach described in~\cite{PRBonn2}.
Next, points belonging to the same object are first clustered together using the angular resolution of the range image $\alpha$, depth disparity, and incidence angle difference $\beta$, similar to the approach proposed in \cite{PRBonn1}; by clustering together points having $\beta$ value larger than a threshold $\beta_{min}$. The identified clusters are then refined by checking for intensity consistency, with the intuition being that LiDAR returns from the same object return similar intensity values. Algorithm \ref{alg:clustering} summarizes the proposed method used for point clustering and object segmentation. It can be noted from lines 17-19 that a neighboring point must satisfy three conditions in order to be added to the same cluster. First, the point should satisfy the depth angle threshold $\beta_{min}$ as discussed above. Second, the intensity of the point must satisfy a minimum threshold $I_{min}$. Third, the intensity of the point must be within a certain intensity range $\pm I_n$ with respect to its neighbors.
\begin{algorithm}
\caption{Image Labeling}\label{alg:clustering}
\begin{algorithmic}[1]
\Procedure{LabelRangeImage}{}
\State \texttt{Label} $\gets 2$, $R \gets$ range img, $I \gets$ intensity img
\State $L \gets zeros(R_{rows} \times R_{cols})$
 \For{ $r = 1...R_{rows}$}
 \For{ $c = 1...R_{cols}$}
 \If{$L(r,c) = 0 $}
 \State \texttt{LabelComponentBFS(r,c,Label)}
 \State \texttt{Label $\gets$ Label$+ 1$}
 \EndIf
\EndFor
\EndFor
\EndProcedure

\Procedure{LabelComponentBFS}{$r,c,$\texttt{Label}}
\State \texttt{queue.push(\{$r,c$\})}
\While{ \texttt{queue} is not empty}
\State \{$r,c$\} $\gets$ \texttt{queue.top()}
\State $L(r,c) \gets$ \texttt{Label}
  \For{ \{$r_n,c_n$\} $\in$ \texttt{Neigborhood\{$r,c$\}}}
  \State $d_1 \gets max(R(r,c), R(r_n,c_n))$
  \State $d_2 \gets min(R(r,c), R(r_n,c_n))$
  \If{$\arctan\frac{d_2\sin\alpha}{d_1 - d_2\cos\alpha} > \beta_{min}$}
  \If{$I(r_n,c_n) > I_{min}$}
  \If{$ \mid I(r_n,c_n) - I(r,c) \mid < I_n$}
  \State \texttt{queue.push(\{$r_n,c_n$\})}
  \EndIf
    \Else
  \State $L(r_n,c_n) \gets$ 1 \Comment{background label}
  \EndIf
  \EndIf
\EndFor
\State \texttt{queue.pop()}
\EndWhile
\EndProcedure
\end{algorithmic}
\end{algorithm}

\begin{figure}
    \centering
    \includegraphics[width=\columnwidth]{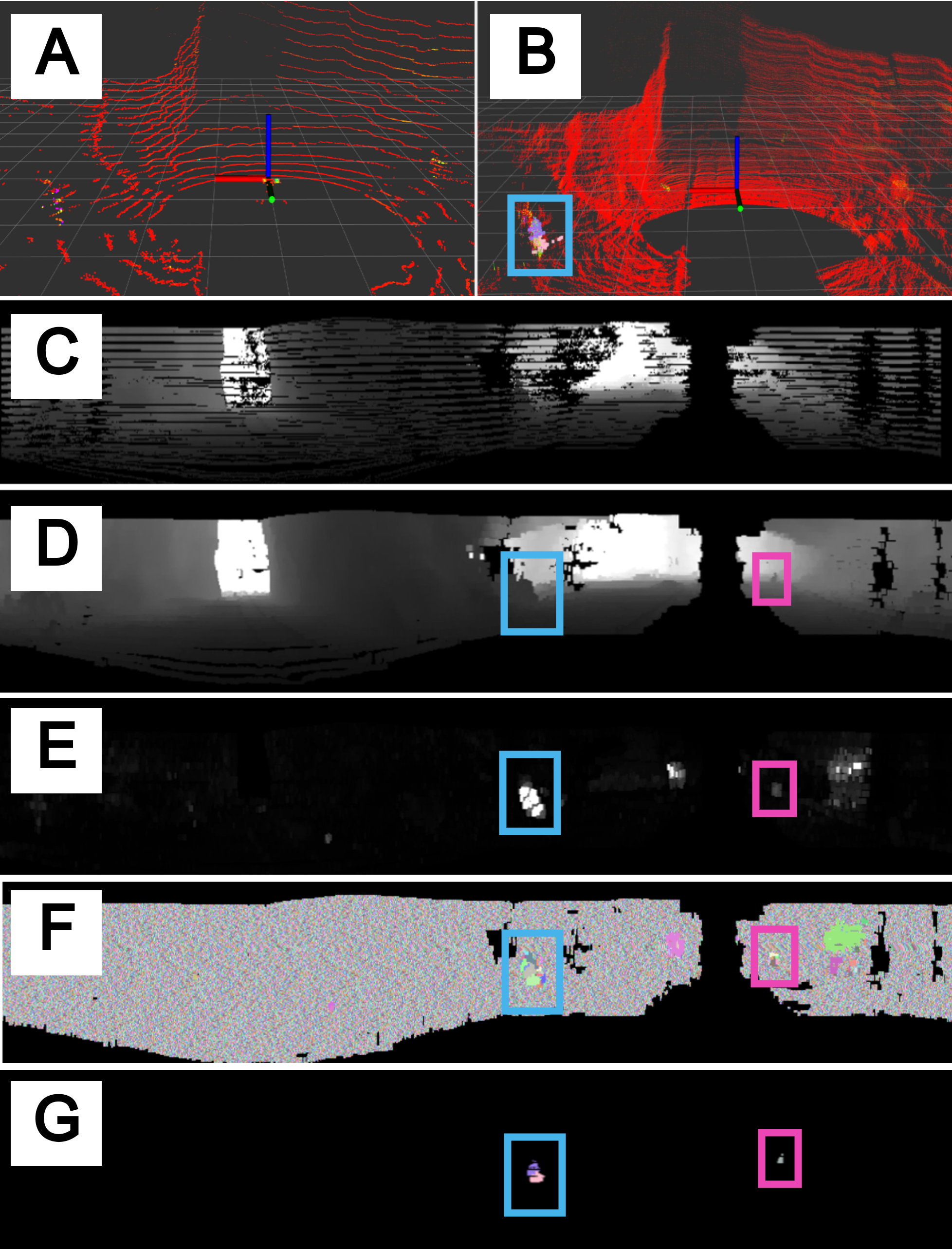}
    \caption{Visualization of the output of each stage of the proposed method. Sub-figures show the single point cloud obtained from VLP-16 LiDAR (A), the accumulated point cloud (B), the range image generated from point cloud projection (C), the filtered range image obtained after bi-linear interpolation and Gaussian blur (D), the corresponding intensity image (E), the segmented object point clusters (F), and the final result obtained after applying various object cluster filters (G). Colored bounding boxes have been added to indicate different objects of interest detected and establish correspondence between sub-figures for better understanding.}
    \label{fig:images}
\vspace{-3ex}
\end{figure}

Once object clusters have been formed, the next step is to remove the object clusters which do not belong to the predefined objects of interest for the mission. Utilizing loose prior knowledge about object properties, such as the range of acceptable size and volume, the clusters are filtered. A cluster is considered valid if it has a volume between $V_{min}$ and $V_{max}$, and the number of points in the cluster is within a range of [$n_{min}$, $n_{max}$]. Moreover, the surface normal standard deviation of the cluster points must lie above a certain threshold \(\sigma_{min}\). The clusters before and after the filtering step are shown in Figure~\ref{fig:images}F and~\ref{fig:images}G, respectively. If multiple clusters lie nearby such that they can be jointly observed with the field-of-view of the used camera, they are combined (cluster merging) to minimize the number of waypoints generated for object candidates to be observed by the camera.



\subsection{Object Proposal Waypoint Generation}
Once an object cluster has been identified, a pose waypoint is generated for the articulated PTZ camera to observe the object cluster and perform object detection. The distance between the object cluster and the camera origin is used to adjust the required zoom level for object observation. Furthermore, to keep track of object clusters that are already observed, a voxel-based global map of the environment is maintained using Voxblox~\cite{oleynikova2017voxblox}. In addition to spatial information, each voxel contains a temporal uniqueness boolean and a timestamp variable to indicate if the camera observed the voxel in the past.
If the number of unobserved voxels in the field-of-view of the PTZ camera is above a certain threshold, the cluster origin in the LiDAR frame is used to generate yaw and tilt angle commands for the PTZ camera in the camera frame for observation. Finally, to detect and classify the observed objects, the learned architecture described in~\cite{dang2020autonomous} is applied to the acquired image from the PTZ camera.

\section{Experimental Results}\label{sec:evaluation}
\subsection{Experimental Setup}
The suitability of the proposed method for real-world disaster response applications was evaluated by collecting sensor data during multiple autonomous field deployments conducted in various underground environments. During experiments, an ANYmal quadruped robot~\cite{hutter2016anymal}, equipped with a Velodyne VLP-16 LiDAR, a PTZ camera system and four static cameras covering the forward hemisphere of the robot, was used. Furthermore, each camera was equipped with an LED flashlight aligned with the camera axis to provide illumination in visually-degraded underground environments. 
It should be noted that this work's main contribution is generating object presence proposals and not object classification. For object proposals generation, LiDAR data was used and the proposals were
compared against classification results and the detection range of static cameras.
For mission autonomy, the robot relied on CompSLAM~\cite{khattak2020complementary} for localization and mapping, with high-level exploration planning provided by GBPlanner~\cite{dang2020graph}. Further details about the hardware and software architecture are provided in~\cite{cerberus}.

To evaluate the object presence proposal performance in complex underground environments, various objects associated with search-and-rescue missions,  such as backpacks, human survivors, drills, helmets, etc., were placed in the environment during field tests. The objects of interest were selected following the "artifacts" specifications provided by the DARPA SubT challenge guidelines\footnote{\href{https://www.subtchallenge.com/resources.html}{https://www.subtchallenge.com/resources.html}}. To evaluate the accuracy of proposals, each unique object proposal generated by the proposed approach was manually evaluated by a human expert, and labeled as an artifact (object-of-interest), non-artifact (an object but not of interest such as wooden boxes, dustbins, water-dispenser, traffic lights, etc., plenty of which were present in DARPA SubT Finals environment) or a false positive (part of environment incorrectly proposed to be an object, e.g. rocks, walls, etc). The precision, defined by the ratio of true object proposals to the total number of proposals, was reported for the experiments, where both artifact and non-artifact detections were considered as true object proposals. Furthermore, the range of detection when using the proposed method against static camera object detections was evaluated by measuring distances for each detection in the robot map of the environment. The set of parameters used for the evaluations are presented in Table~\ref{tab:params}. 
\begin{table}[h]
\centering
\caption{Parameter values used during evaluation.}
\begin{tabular}{l|c|l|c}
\hline
\multicolumn{1}{c|}{Parameter} & Value     & \multicolumn{1}{c|}{Parameter} & Value          \\ \hline
\(N_c\)               & 10        & \(I_{min}\)                      & 25             \\
\(\theta_{min}\)($\degree$)      & 30        & \(I_n\)                         & 60             \\
\(d_{min} (\si{\meter})\)       & 0.15   & \(V_{min}, V_{max}\)(\si{\meter^{3}})                 & 0.01, 0.8  \\
\(f (Hz)\)            & 2       & \(n_{min}, n_{max}\)                 & 50, 5000       \\
\(row, col\)          & 180, 1200 & \(\sigma_{max}\)                       & 0.01           \\
\(V_{FOV}\)($\degree$)          & 60 & \(\beta\) ($\degree$)             & 14        \\ \hline
\end{tabular}
\label{tab:params}  
\vspace{-3ex}
\end{table}

\begin{figure*}
    \centering
    \includegraphics[width=\textwidth]{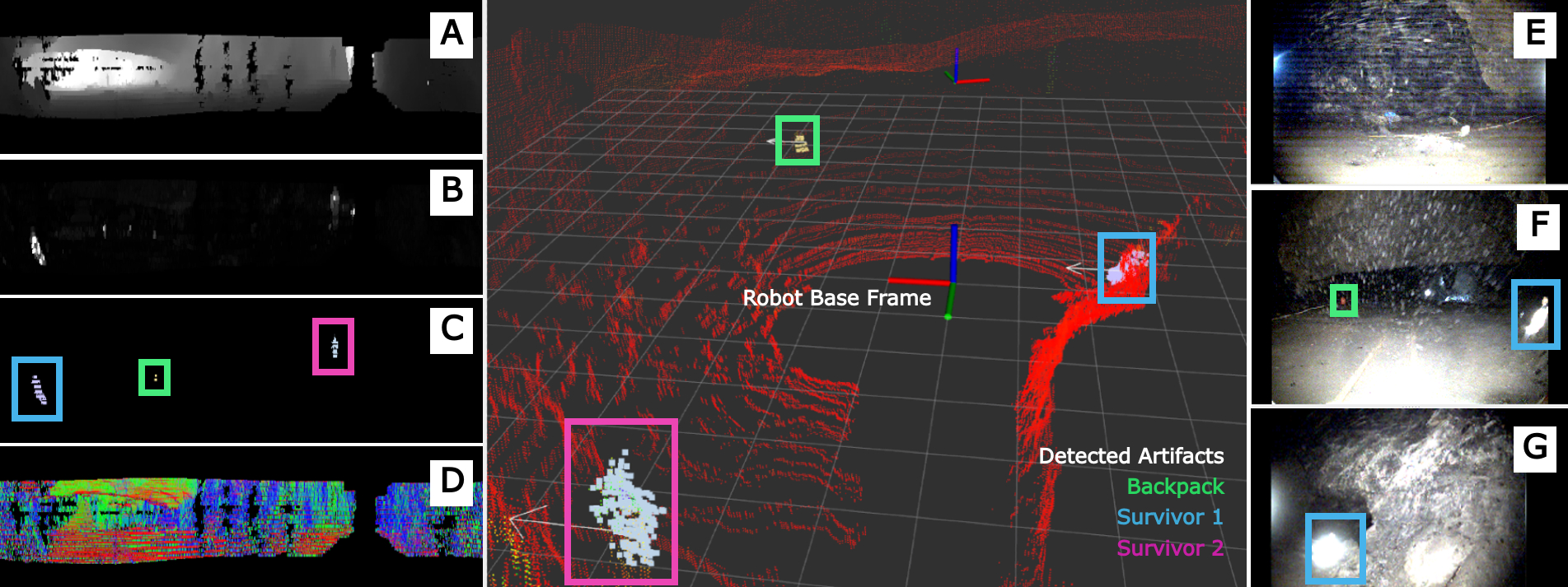}
    \caption{An instance during the Seemuhle underground mine experiment using the proposed method to detect three artifacts. Sub-figures show the filtered range image (A), the filtered intensity image (B), the segmented object points clusters (C), the surface normal image (D), images from left (E), center (F), and right (G) static cameras for reference. The colored bounding boxes correspond to the same detected artifacts across images and are visualized to provide a better understanding.}
    \label{fig:rviz}
\vspace{-3ex}
\end{figure*}

\subsection{Seemuhle Underground Mine Deployment}\label{sec:seemuhle}
For evaluation, an autonomous robot exploration mission was conducted at Seemühle Borner Walenstadt mine in Switzerland under the conditions of complete darkness. During this mission, a total of 34 unique object proposals were generated and provided to the PTZ camera for classification, with a precision of 85.3\%. To provide a qualitative understanding, Figure~\ref{fig:rviz} shows the output of each sub-module of the proposed work for an instance during the experiment when three artifacts (two human survivors and a backpack) were detected to be in the vicinity of the robot. Quantitative results comparing the object detection distance of the proposed approach against using static cameras are summarized in Table~\ref{tab:seemuehle}. If multiple instances of the artifact are detected, a detection distance range instead of a single value is provided. It can be clearly noted that the proposed method's detection ranges extend those of the static cameras. It can be argued that the overall detection ranges are not very large for both methods; however, the maximum observation ranges are limited due to the complex topology of the underground mine. This is clearly demonstrated when multiple instances of the same artifact are present. The proposed method can detect the artifact when it is first observable, thus providing multiple detection distances, whereas the static cameras can only detect them at a fixed distance when the robot is very near to the object. It can also be noted that during this experiment our method was not able to detect the rope artifact. This can be attributed to the small size and relatively low-intensity LiDAR returns obtained from this specific artifact. Nevertheless, the robot is still able to detect the object at a closer distance using the static cameras showcasing the usefulness of our approach to augment the current artifact detection pipeline.

\begin{figure}
    \centering
    \includegraphics[width=\columnwidth]{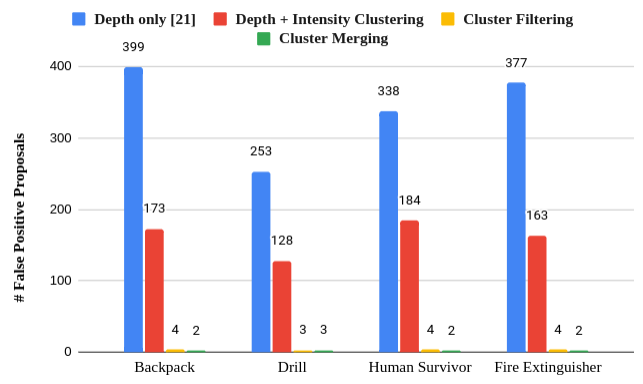}
    \caption{Comparison of the number of false positive proposals generated for different object point cluster segmentation methods and strategies for three types of artifacts detected during the Seemuhle underground mine mission.}
    \label{fig:False_positives}
\vspace{-3ex}
\end{figure}

To demonstrate the advantage of the proposed method over the current state-of-the-art LiDAR depth-only object segmentation methods, a comparison with the approach proposed in~\cite{PRBonn1} is performed. Furthermore, to demonstrate the effectiveness of utilizing auxiliary LiDAR data (intensity returns) and the proposed filtering steps, an ablation study is conducted. The results for the comparison and ablation study are presented in Figure~\ref{fig:False_positives}, which compares the number of false positive (FP) proposals for four types of artifacts detected over a horizon of $100$ point clouds. For this study, four short sequences of the dataset were used wherein only one artifact was present near the robot for ease of calculating the FPs associated with the artifacts. The efficacy of the proposed approach and all individual steps is clearly demonstrated by the low number of false positive (FP) proposals reported.

\subsection{DARPA SubT Challenge Finals}\label{sec:darpa}
The DARPA SubT Challenge's Final Event took place at the Louisville Mega Cavern in Kentucky on $21-24$ September \num{2021}. During its hour-long final run, team CERBERUS deployed four ANYmal-C robots to autonomously explore the unknown underground environment containing caves, tunnels, and urban structures, collectively travelling a distance of over ~\SI{1.7}{\km}. For the evaluation of this work, we selected the longest robot trajectory and a total of 334 unique object proposals were generated, with a precision of 53.4\%. It can be noted that compared to the previous result, the precision of object proposals is lower. One probable reason is the inconsistency of LiDAR intensity returns obtained in this artificially created finals course compared to those obtained in the natural environment of the previous experiment. Nevertheless, even in such unfavourable conditions, the proposed method is able to demonstrate its efficacy by consistently detecting objects at a farther distance than the static cameras, as shown quantitatively in Table \ref{tab:seemuehle}.

\begin{table}[]
\centering
\caption{Object detection range (\si{\meter}) comparison}
\begin{tabular}{l|l|c|c}
\hline
\multicolumn{1}{c|}{\textbf{Dataset}} & \multicolumn{1}{c|}{\textbf{Artifact Name}} & \textbf{\begin{tabular}[c]{@{}c@{}}Proposed  \\ Method\end{tabular}} & \textbf{\begin{tabular}[c]{@{}c@{}}Static \\ Camera\end{tabular}} \\ \hline
\multirow{5}{*}{Seemuehle}            & Backpack                                    & 5-10                                                                 & 3.5                                                                       \\
                                      & Human Survivor                              & 4-9                                                                  & 3                                                                         \\
                                      & Power Drill                                 & 3.5-6.5                                                              & 5                                                                         \\
                                      & Rope                                        & -                                                                    & 2.5                                                                       \\
                                      & Fire Extinguisher                           & 4.5-6                                                                & 3.5                                                                       \\ \hline
\multirow{4}{*}{DARPA}                & Backpack                         & 4.5-13                                                               & 3.5                                                                       \\
                                      & Human Survivor                         & 2.5                                                                  & 2.5                                                                       \\
                                      & Power Drill                           & 4.5                                                                  & 2.5                                                                       \\
                                      & Vent                             & 3.5-5                                                                & 1.5                                                                       \\ \hline
\end{tabular}
\label{tab:seemuehle}
\vspace{-3ex}
\end{table}

\section{Discussion \& Future Work}\label{sec:conclusions}
\label{sec:conclusion}
In this work, we presented an approach that utilized multi-modal sensing to improve the range of object detection in complex and visually-degraded underground environments. We demonstrated that exploiting auxiliary intensity data in combination with depth data can improve distinguishing objects from the environment structure and hence aid in creating detection proposals for a camera system. The proposed work, tested on data collected in complex environments, showed that it can improve the object detection range without reducing mission speed by not requiring continuous scanning for full coverage from an articulated camera system.
We observed that the proposed method worked better in natural underground environments like caves and mines, where the homogeneous background environment returns consistent intensity values compared to artificial man-made environments, making it feasible to distinguish objects from the environment. Although we tested LiDAR intensity calibration techniques~\cite{intensitycalib}, the performance improvement was minimal. In future, we aim to study different LiDAR sensors and develop a generalized intensity calibration method. Furthermore, we aim to tightly couple the object proposal generation with the path planning stack to perform an active object search for time-critical search-and-rescue missions.

\bibliographystyle{IEEEtran}
\bibliography{0_main}

\end{document}